\documentclass[10pt,twocolumn,letterpaper]{article}
\usepackage{cite}
\usepackage{cvpr}
\usepackage{times}
\usepackage{epsfig}

\usepackage{amsmath}
\usepackage{amssymb}
\usepackage{paralist}
\usepackage{color}
\usepackage{float}
\usepackage{comment}
\usepackage{multirow}
\usepackage{enumitem}
\usepackage{subfigure}
\usepackage{url}

\usepackage[pagebackref=true,breaklinks=true,letterpaper=true,colorlinks,bookmarks=false]{hyperref}

\cvprfinalcopy 

\def\cvprPaperID{5803} 

\def\eg{\emph{e.g}\onedot} 
\def\ie{\emph{i.e}\onedot} 
 
\def\etc{\emph{etc}\onedot} 
 
\def\etal{\emph{et al}\onedot}
\begin{document}

\title{SR-LSTM: State Refinement for LSTM \\
towards Pedestrian Trajectory Prediction}

\author{Pu Zhang{\small $~^{1}$}, ~Wanli Ouyang{\small $~^{2}$}, Pengfei Zhang{\small $~^{1}$}, ~Jianru Xue{\small $~^{1}$}, ~Nanning Zheng{\small $~^{1}$}\\
	\normalsize
	$^{1}$\
Institute of Artificial Intelligence and Robotics, Xi¡¯an Jiaotong University, China\\
	\normalsize
$^{2}$\, The University of Sydney, SenseTime Computer Vision Research Group, Australia\\
	\normalsize
	\normalsize
	{zhangpu2016,zpengfei}@stu.xjtu.edu.cn,\\
	\normalsize
	{jrxue,nnzheng}@mail.xjtu.edu.cn,\\
	\normalsize
    wanli.ouyang@sydney.edu.au\\
}
\maketitle

\begin{abstract}
In crowd scenarios, reliable trajectory prediction of pedestrians requires insightful understanding of their social behaviors. These behaviors have been well investigated by plenty of studies, while it is hard to be fully expressed by hand-craft rules. Recent studies based on LSTM networks have shown great ability to learn social behaviors. However, many of these methods  rely on previous neighboring hidden states but ignore the important current intention of the neighbors. In order to address this issue, we propose a data-driven state refinement module for LSTM network (SR-LSTM), which activates the utilization of the current intention of neighbors, and jointly and iteratively refines the current states of all participants in the crowd through a message passing mechanism. To effectively extract the social effect of neighbors, we further introduce a social-aware information selection mechanism consisting of an element-wise motion gate and a pedestrian-wise attention to select useful message from neighboring pedestrians. Experimental results on two public datasets, \ie ETH and UCY, demonstrate the effectiveness of our proposed SR-LSTM and we achieves state-of-the-art results.

\end{abstract}

\section{Introduction}\label{sec:intro}

Pedestrian trajectory prediction is strongly required by various applications, \eg, autonomous driving and robot navigation. The trajectory of pedestrian can be influenced by multiple factors such as scene topologies, pedestrian beliefs, and the most complex one, human-human interactions. The intricate and subtle interactions are often taken place among the pedestrians. For example, strangers avoid collisions, but fellows walk in group. Broken groups can regroup to keep the unity\cite{mccool2017simulating,gorrini2015empirical}. When individuals meet groups, singles are statistically walking faster and are more likely to adjust their routes\cite{do2016group,gorrini2016age}. Stationary groups act as obstacles\cite{yi2015understanding, yi2014profiling}.

Although various social behaviors have been investigated, it is challenging to take a comprehensive consideration of them.
Some recent data-driven methods\cite{varshneya2017human,vemula2018social, hasan2018mx, alahi2016social, gupta2018social, sadeghian2018sophie,xu2018encoding, su2017forecast} try to leverage from Long-Short Term Memory networks (LSTM)\cite{hochreiter1997long}, to learn social behaviors from large scale data. In this paper, we point out two factors which are important but neglected in different levels:
\begin{figure}[t]
\begin{center}
\includegraphics[width=0.95\linewidth]{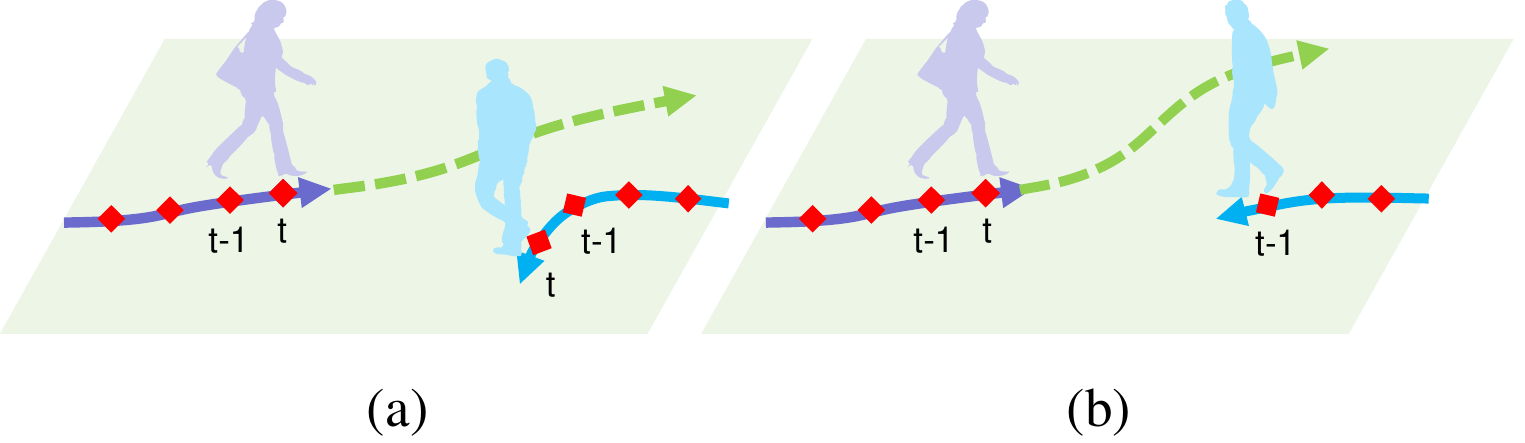}

\caption{When predicting for the lady at time $t$, considering the trajectory of the man on the right up to time $t$ (a), or the one up to time $t-1$ (b), can cause great deviation in predicting results (dashed lines).}
\end{center}
\label{fig:motivation1}
\end{figure}

1). \emph{Current states of neighbors are important for timely interaction inference.}\par
Many of the recent RNN-based approaches make use of the previous hidden states of neighbors \cite{alahi2016social,sadeghian2018sophie, gupta2018social, hasan2018mx, varshneya2017human, su2017forecast}. However, the previous states fail to reveal the newest status of neighbors especially when they have just change their intentions in short time period. This effect of lagging depends on the size of the time step. Within a common time step in recent works\cite{alahi2016social,gupta2018social,sadeghian2018sophie}, \eg, 0.4s, human can take one stride, in which the intentions of them could change unexpectedly. Fig. \ref{fig:motivation1} shows an example. The man on the right in Fig. \ref{fig:motivation1}(a) changes his intention to turn left at time $t$. Based on this observation, the predictions of the lady can be straight on or turning slightly. But if the algorithm only considers the neighbors' trajectory till $t-1$ (Fig. \ref{fig:motivation1}(b)), the man tends to go straight and forces the lady to largely turn and avoid collision, which results in large prediction error. 
Therefore, we are motivated to take advantage of the current neighboring states into consideration.

2). \emph{Useful information should be adaptively selected from neighbors, based on their motions and locations.}\par
\begin{figure}[t]
\begin{center}
\subfigure[]{
\includegraphics[width=0.4\linewidth]{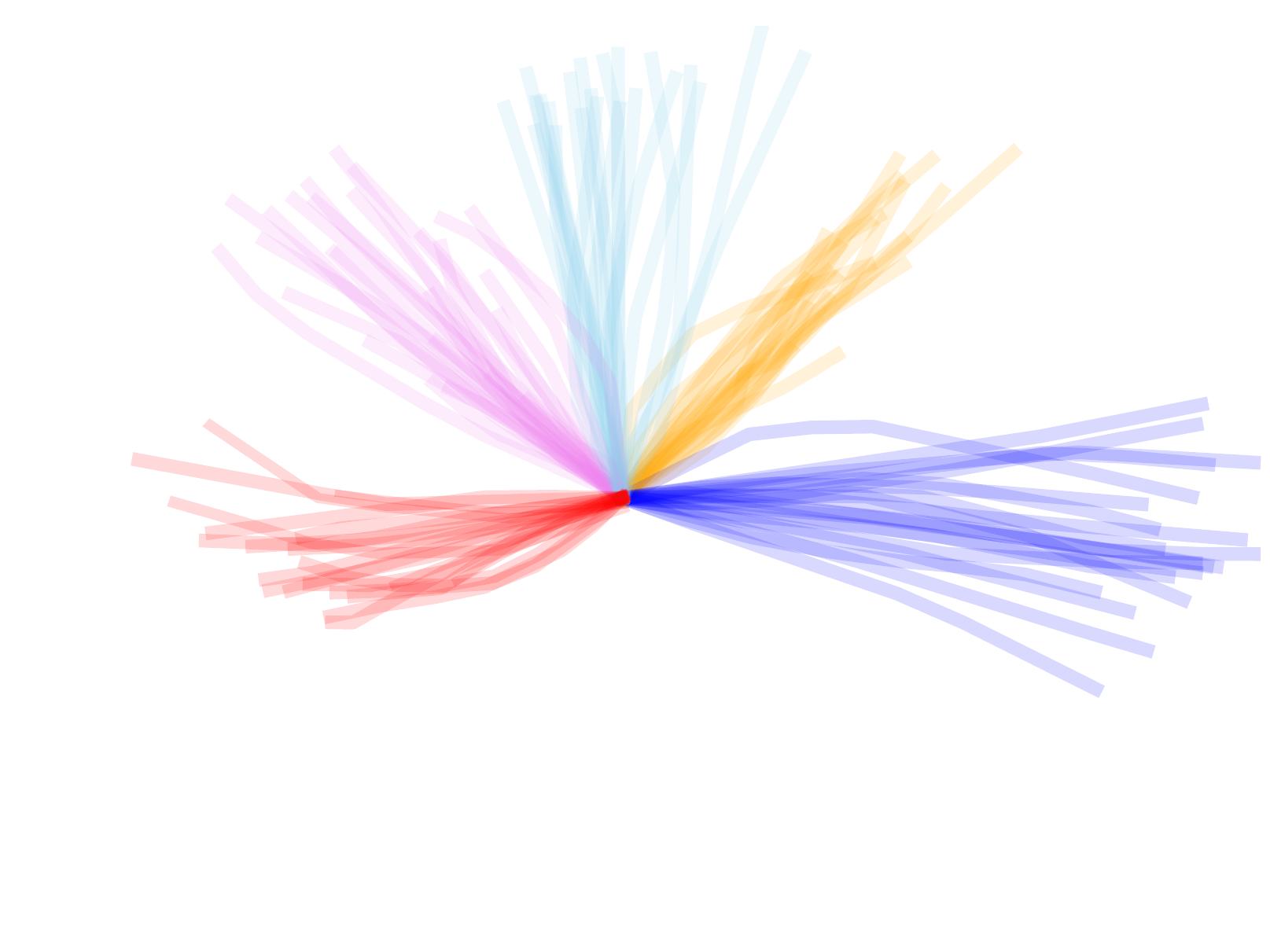}\label{fig:hidden_pattern}}
\hspace{0.15in}
\subfigure[]{
\includegraphics[width=0.5\linewidth]{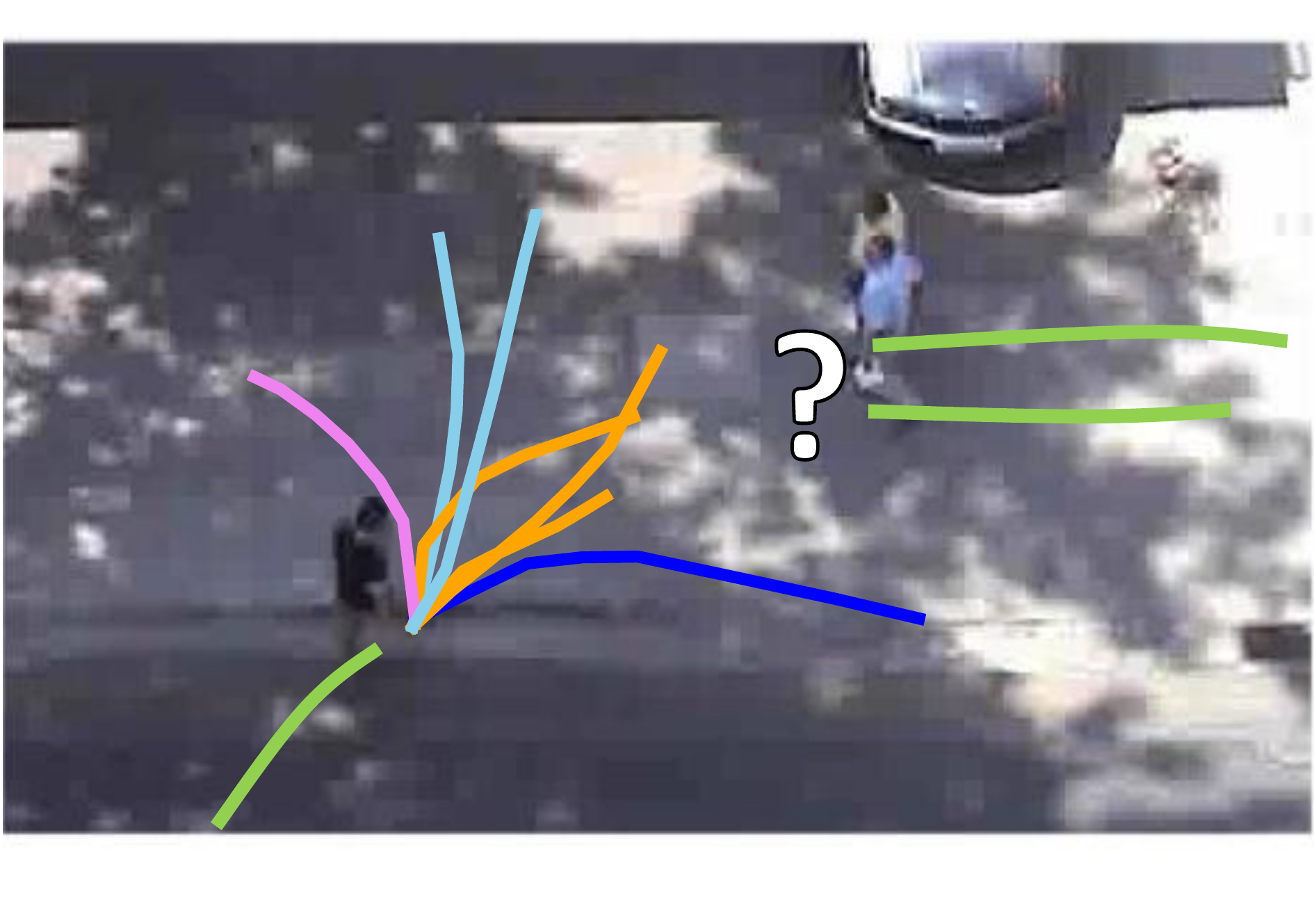}\label{fig:collisionavoid}}
\end{center}
\caption{(a) Activation trajectory patterns of hidden neurons in LSTM, which start from the origin. 
Each trajectory pattern marked by certain color contains trajectories from database which has top-20 responses for the hidden neuron. (b) A sample of three pedestrian interaction. How will the dyad pay attention to the other pedestrian on the left?} 
\label{fig:motivation2}
\end{figure}

Neural networks, \eg, LSTM, can be used for extracting the features representing the trajectory.
To better explain these features, Fig. \ref{fig:hidden_pattern} visualizes the trajectory pattern captured by each feature in the LSTM. It can be seen that these neurons are responsible for various motion patterns covering the walking direction and speed. Many approaches utilize the features of neighboring pedestrians to estimate the trajectory of a pedestrian.
However, the features (motion patterns) of neighboring pedestrians are not equally important for predicting the trajectory of a pedestrian. As shown in Fig. \ref{fig:collisionavoid}, the two pedestrians on the right mostly pay more attention to the situation of collision, which can be represented by the trajectory features of the other pedestrian on the left walking towards them. This potential attention depends on the pairwise motion and relative location of the pedestrian to be predicted and his neighbor. Notably, each neighbor should be particularly treated because different kinds of attention should be assigned to pedestrians according to different interaction conditions.
Based on this motivation, we introduce a motion gate to select the most
useful features from
each neighbor, based on the pairwise motion character and relative location.

In this paper, we propose a states refinement module for LSTM (SR-LSTM), which aligns all pedestrians together and adaptively refine the state of each participant through a message passing framework. Further more, the refinement process can be performed for multiple times, indicating deeper interactions among the crowds. SR-LSTM focuses on the adjustment for current LSTM states, which is quite different from existing RNN-based approaches. To adaptively extract social effects from neighbors for feature refinement, we further introduce a social-aware information selection mechanism, consisting of an element-wise motion gate and a pedestrian-wise attention layer.

Contributions of this paper are summarized as follows:
\begin{itemize}[noitemsep,topsep=0pt,parsep=0pt,partopsep=0pt]
\item
A novel interactive recurrent structure, SR-LSTM, is proposed as a new pipeline for jointly predicting the future trajectories of pedestrians in the crowd.
\item
SR-LSTM aligns all pedestrians in the scene to adaptively refine the current states of each other. The refinement can be performed for multiple times to model the deep interactions between humans.
\item
Motion gate is introduced to effectively focus on the most useful neighborhood features.
\end{itemize}
\section{Related Work}

\textbf{Research on human-human interaction.}
Early work from Helbing and Molnar\cite{helbing1995social} models the interaction between humans as ``social force", which is proved to be effective and applied to crowd analysis\cite{mehran2009abnormal,helbing2005self} and robotics\cite{ratsamee2013human,ferrer2013robot}. Succeeding methods take more potential factors into account, such as pedestrian attributes\cite{zhang2012abnormal,yi2015understanding}, walking group\cite{moussaid2010walking,yamaguchi2011you}, stationary group\cite{yi2015understanding, yi2014profiling}. Some studies based on game theory model the interaction among pedestrian flows\cite{yu2010game,hoogendoorn2003simulation} and evacuation process\cite{zheng2011conflict,hoogendoorn2013modeling,bouzat2014game}, Ma \textit{et al.}\cite{ma2017forecasting} predict pedestrians from a static frame using fictitious play. Most of these methods are based on hand-craft functions and rules, which might fail to generalize for more complex interaction cases.

\textbf{RNNs based approaches for trajectory prediction.}
Recently, Recurrent Neural Networks (RNN) and its variant structures such as
LSTM\cite{hochreiter1997long} and Gated Recurrent
Units (GRU)\cite{chung2014empirical} are widely used in various tasks including pedestrian trajectory forecasting\cite{varshneya2017human,vemula2018social, hasan2018mx, lee2017desire, alahi2016social, gupta2018social, sadeghian2018sophie, su2017forecast,su2016crowd,xu2018encoding}, where each pedestrian is modeled by RNN with shared parameters.
In order to model the human interactions, researchers follow two primary ways to involve information of neighbors, using their current observations \cite{vemula2018social, gupta2018social, sadeghian2018sophie, lee2017desire, su2017forecast} (such as velocity, location, \etc.) or introducing previous states into current RNN recursion\cite{alahi2016social, lee2017desire, sadeghian2018sophie, gupta2018social, hasan2018mx, varshneya2017human, su2017forecast,su2016crowd}.
These methods treat the information of neighboring pedestrians as input which serves in an input-to-output mechanism. 
In comparison, we treat the information from neighboring pedestrians as message provider and construct a message passing mechanism to refine the features of each other. Therefore, our approach uses the information from current time step and can refine the information through multiple message passing iterations.

\textbf{Attention based approaches for trajectory prediction.}
Attention mechanisms have been proven to be significantly effective for relevant data selection in various tasks\cite{vaswani2017attention,xu2015show,wang2016hierarchical,li2017attentive}.
Some RNN-based works for pedestrian trajectory prediction utilize the attention mechanism to distinguish the importance of different neighbors \cite{vemula2018social,fernando2018soft+,sadeghian2018sophie,su2016crowd,su2017forecast}. Vemula \etal\cite{vemula2018social} compute a soft attention score from the hidden states of the designed edgeRNNs, which gives an importance value for each neighbor. Sadeghian \etal\cite{sadeghian2018sophie} utilize the soft attention similar with \cite{xu2015show} to highlight the important neighbors. 
Su \etal \cite{su2016crowd, su2017forecast} calculate the pairwise velocity correlation, and emphasis the neighbors who are in similar velocity.
However, our motion gate aims to selects motion features from each neighboring pedestrian during the refinement, which can extract more socially aware neighboring features and has not been employed in previous approaches..

\textbf{Graph-based and message passing framework.}
This work is also inspired by Graph Convolution Networks (GCN)\cite{bruna2013spectral,kipf2016semi} and message passing frameworks used for other applications such as object detection\cite{hu2018relation,yuan2017temporal}, action recognition\cite{yan2018spatial,shi2018adaptive}, semantic segmentation\cite{liang2016semantic,liang2017interpretable}, scene graph generation\cite{xu2017scene,yang2018graph,li2017scene,li2017vip}, video recognition\cite{wang2018videos}, \etc.

Our method treats the pedestrian walking space as a fully connected graph and which can be regarded as a variant of GCN specially designed for the trajectory prediction task. We consider message passing for pedestrians within constrained regions, and use pairwise motion character and relative spatial location between pedestrians for guiding message passing.

\section{Method}

\textbf{Problem formulation }
In this paper, we address the problem of pedestrian trajectory prediction in the crowd scenes. We focus on the two-dimensional spatial coordinates at specific time intervals. 
For the given observed trajectories including $T_{obs}$ frames and $N$  pedestrians, the trajectory point of the $i$th pedestrian on the $t$th frame is represented by $(x_i^t,y_i^t)$.
The problem is defined to predict the future trajectories $(\hat{x}_i^t,\hat{y}_i^t)$, where $t=T_{obs}+1,T_{obs}+2, \ldots$ 

\subsection{Vanilla LSTM}\label{sec:vlstm}
Vanilla LSTM (V-LSTM) model infers all pedestrian independently, without considering the interactions among them.
At time $t$, the location of the $i$th pedestrian is embedded as a vector $e_i^t = \phi_e(x_i^t, y_i^t; W_{e})$, where $\phi_e$ is the embedding function parameterized by $W_e$. 
The vector $e_i^t$ is used as the input to the LSTM cell as follows:
\begin{equation}
\begin{aligned}\label{eq:vlstm}
g_i^{u,t} &=\delta(W^ue_i^t+U^uh_i^{t-1}+b^u),\\
g_i^{f,t} &=\delta(W^fe_i^t+U^fh_i^{t-1}+b^f),\\
g_i^{o,t} &=\delta(W^oe_i^t+U^oh_i^{t-1}+b^o),\\
g_i^{c,t} &=\mathrm{tanh}(W^ce_i^t+U^ch_i^{t-1}+b^c),\\
c_i^{t} &=g_i^{f,t}\odot c_i^{t-1}+g_i^{u,t}\odot g_i^{c,t},\\
h_i^{t} &=g_i^{o,t}\odot \mathrm{tanh}(c_i^{t}),\\
\end{aligned}
\end{equation}
where $g$ denotes the gate function inside the LSTM cell, the superscripts $u$,$f$,$o$,and $c$ denote the update gate, forget gate, output gate and cell gate, respectively. $W$ and $U$ denote the weight matrix connecting input and hidden state to the LSTM cell. A pedestrian will be treated as a sample when using LSTM. All LSTM parameters are shared across pedestrians.

With the hidden states $h_i^t$ extracted from LSTM, we directly predict the coordinates at time step $t+1$ following \cite{gupta2018social}:
\begin{equation}
\begin{aligned}\label{eq:pred}
[\hat{x}_i^{t+1}, \hat{y}_i^{t+1}]^\textrm{T} =W_p h_i^{t}, \\
\end{aligned}
\end{equation}
where $W_p$ is the learned parameter. The parameters of the LSTM model are directly learned by minimizing
the L2 loss between predicted position and ground truth.
In the inference stage, the coordinates predicted from the previous time step are used as the input at the current time step.

\subsection{The SR-LSTM Framework}

\begin{figure}[t]
\begin{center}

\includegraphics[width=0.9\linewidth]{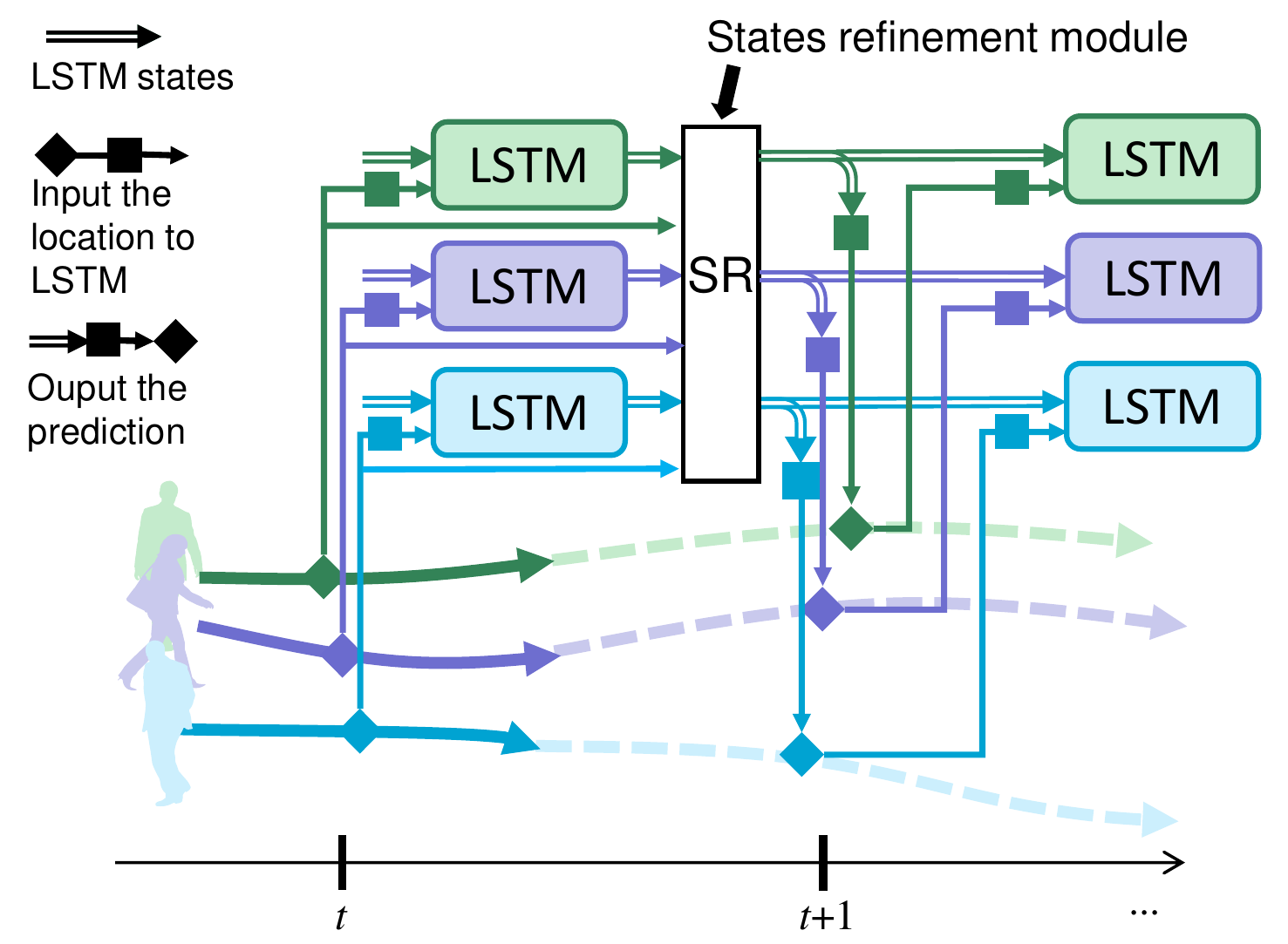}
\end{center}
\caption{Framework overview of proposed SR-LSTM. States refinement module is considered as an additional subnetwork of the LSTM cells, which aligns pedestrians together and updates current states of them. The refined states are used to predict the location at the next time step.}
\label{fig:overview}
\end{figure}

The overview of SR-LSTM framework is illustrated in Fig. \ref{fig:overview}. In this framework, the LSTM in Section \ref{sec:vlstm} is used for extracting features from the trajectory of each pedestrian separately. The main difference is that the States Refinement (SR) module is used for refining the \ie cell states $c_i^{t}$ in Eq. \ref{eq:vlstm} by passing message among pedestrians.

The SR module takes the following three information sources of all pedestrians as input: the current locations of pedestrians, hidden states and cell states from LSTM. The output of the SR module is the refined cell states.
Mathematically, the SR module for refining the cell states can be formulated as follows:
\begin{equation}
\begin{aligned}
\hat{c}_i^{t,l+1}&=\sum_{j\in \mathcal{N}(i)}M_{j}(\hat{h}_j^{t,l}, \hat{h}_i^{t,l})+\hat{c}_i^{t,l}, \\
\end{aligned}\label{eq:Mp}
\end{equation}
where $M$ is the message passing function detailed in Section \ref{sec:selection}.  $\mathcal{N}(i)$ denotes the neighbors of pedestrian $i$.
For the $i$th pedestrian, the hidden states $\hat{h}_j^{t,l}$ from neighboring pedestrians for $j\in \mathcal{N}(i)$ are integrated through the function $M$ and then combined with the cell state of $i$ to obtain the refined cell state.
Message passing can be done for multiple times. $l$ denotes the message passing iteration index.
The states with $l=0$ are initialized by the original LSTM states in Eq. \ref{eq:vlstm}.

After the cell states are refined by $L$ refinement iterations in the SR module,
they are used for producing predicted coordinates as follows:
\begin{equation}
\begin{aligned}
\hat{c}_j^{t}=c_j^{t,L}.\\
\hat h_i^{t} =g_i^{o,t}\odot \mathrm{tanh}(\hat c_i^{t}),\\
\end{aligned}\label{eq:finalstates}
\end{equation}
\begin{equation}
\begin{aligned}\label{eq:SRpred}
[\hat{x}_i^{t+1}, \hat{y}_i^{t+1}]^\textrm{T} =W_p \hat{h}_i^{t}, \\
\end{aligned}
\end{equation}
where $g_i^{o,t}$ is from the LSTM.
In the task of pedestrian trajectory prediction, further refinement could improve the quality of the interaction model, indicating the intention negotiation of human interaction natures.

\subsubsection{A simple implementation of message passing}
A simple implementation of message passing can be formulated as follows:
\begin{equation}
\begin{aligned}
\hat{c}_i^{t,l+1} &=\sum\limits_{j\in \mathcal{N}(i)}W^{mp}\hat{h}_j^{t,l}/|\mathcal{N}_i|+\hat{c}_i^{t,l},\\
\end{aligned}\label{eq:Mp2}
\end{equation}
where $|\mathcal{N}_i|$ denotes the number of elements in $\mathcal{N}(i)$. Message passing function $M_{j}(\hat{h}_j^{t,l}, \hat{h}_i^{t,l})=W^{mp}\hat{h}_j^{t,l}/|\mathcal{N}(i)|$, which does not depend on $\hat{h}_i^{t,l}$. $W^{mp}$ is a linear transformation using for transmitting the message from neighboring pedestrians to the pedestrian $i$.

When using the features from other pedestrians, treating all their features equally is not an appropriate solution. We  design more effective message passing term $M$ in following section.

\subsubsection{Social-aware information selection }\label{sec:selection}
To adaptively focus on the most useful neighboring information and guide the message passing,  we design the following message passing term $M$ with a social-aware information selection mechanism:
\begin{equation}
\begin{aligned}
\hat{c}_i^{t,l+1}&=\sum_{j\in \mathcal{N}(i)}M_{j}(\hat{h}_j^{t,l}, \hat{h}_i^{t,l})+\hat{c}_i^{t,l},\\
&=\sum\limits_{j\in N(i)}W^{mp}\alpha_{i,j}^{t,l} \cdot (g_{i,j}^{m,t,l}\odot  \hat{h}_j^{t,l})+\hat{c}_i^{t,l},\\
\end{aligned}\label{eq:M}
\end{equation}
where $\odot$ denotes the element-wise product operation. As that in Eq. \ref{eq:Mp2}, $W^{mp}$ is the linear transform parameter. The pedestrian-wise attention $\alpha_{i,j}$ and motion gate $g_{i,j}$ in Eq. \ref{eq:M} are introduced below.

\textbf{Pedestrian-wise attention.} $\alpha_{i,j}$ in Eq. \ref{eq:M} is a scalar.
It is the attention for pedestrian $j$ formulated as follows:
\begin{equation}
\begin{aligned}\label{eq:attention}
u_{i,j}^{t,l}&={w^a}^\textrm{T}[r_{i,j}^{t,l};\hat{h}_j^{t,l};\hat{h}_i^{t,l}],\\
\alpha_{i,j}^{t,l}&=\frac{\mathrm{exp}(u_{i,j}^{t,l})}{\sum\limits_{k}{\mathrm{exp}(u_{i,k}^{t,l})}},\\
\end{aligned}
\end{equation}
where $r_{i,j}^{t,l}$ is the relative spatial location, which is an important factor to guide the information selection. It is embedded by embedding function $\phi_r$ as follows:
\begin{equation}
\begin{aligned}\label{eq:r}
r_{i,j}^{t,l}&=\phi_r(x_{i}^t-x_{j}^t,y_{i}^t-y_{j}^t;W^r),\\
\end{aligned}
\end{equation}
where $(x_{i}^t, y_{i}^t)$ is the location of pedestrian $i$ at time $t$, similarly for $(x_{j}^t, y_{j}^t)$. $W^r$ denotes the parameters for the embedding function $\phi_r$.

\textbf{Motion gate.} $g_{i,j}^m$ is a vector, which is formulated as:
\begin{equation}
\begin{aligned}\label{eq:gm}
g_{i,j}^{m,t,l}&=\delta(W^m[r_{i,j}^{t,l};\hat{h}_j^{t,l};\hat{h}_i^{t,l}]+b^{m}),\\
\end{aligned}
\end{equation}
$g_{i,j}^m$ is designed for feature selection, where $W^m$, $b^m$ are parameters and $\delta$ denotes the sigmoid function. $g_{i,j}^{m,t,l}$ selects features by using the element-wise product $g_{i,j}^{m,t,l}\odot  \hat{h}_j^{t,l}$ in Eq. \ref{eq:M}.

\begin{table*}
\small
\begin{center}
\begin{tabular}{c|ccc|c|c|c|c|c|c}
\hline %
\multirow{2}{*}{ID} &\multicolumn{3}{|c|}{Pre-processing} & \multicolumn{6}{|c}{Performance (MAD/FAD)} \\
\cline{2-10}
&Rela/Nabs &EUf &RR  &ETH-univ  &ETH-hotel &UCY-zara01 &UCY-zara02 &UCY-univ &AVG\\
\hline
\hline
1&Rela&-         &-                    &1.16/2.29 &0.57/1.07 &0.68/1.39 &0.61/1.27 &0.76/1.60 &0.76/1.52\\
2&Nabs&-         &-                   &1.00/2.04 &0.50/1.08 &0.58/1.30 &0.40/0.87 &0.64/1.38 &0.63/1.33\\
3&Nabs&\checkmark&-                     &0.84/1.90 &0.45/0.94 &0.43/0.94 &0.38/0.87 &0.63/1.42 &0.55/1.21\\
4&Nabs&\checkmark&\checkmark         &0.83/1.77 &0.41/0.80 &0.49/1.15 &0.37/0.85 &0.56/1.22 &0.53/1.16\\
\hline
\end{tabular}
\end{center}
\caption{Performance on V-LSTM with different data pre-processings. \textbf{Rela}: differentiate the sequences as relative spatial offsets. \textbf{ Nabs}: use the absolute position but shift the origin to the latest observed time slot. \textbf{EUf}: frame rate correction on ETH-univ. \textbf{RR}: random rotation for each data mini-batch. We adopt the configuration of ID 4 for our experiments.}
\label{table:preprocess}
\end{table*}
The motion gate and the pedestrian-wise attention have different functionalities and jointly select the important information from neighboring pedestrians for message passing. Further explanation of these two components are as follows:
\begin{itemize}[noitemsep,topsep=0pt,parsep=0pt,partopsep=0pt]
\item 
The motion gate $g_{i,j}^m$ acts on each hidden state $\hat{h}_j^{t}$ to perform a pairwise feature selection. It is calculated based on the combination of $r_{i,j}^{t}$, $\hat{h}_j^{t}$, $\hat{h}_i^{t}$ (see Eq. \ref{eq:gm}), which suggests that the motion of pedestrian $i$ and $j$ and their relative spatial location are jointly considered for feature selection. This element-wise feature selection can not be provided by the pedestrian-wise attention. 
\item  The pedestrian-wise attention is to emphasize important neighbors and control the amount of neighborhood message. If we only take the motion gate, training process could hardly converge due to the uncertain number of correlated neighbors.
\item The simple implementation in Eq. \ref{eq:Mp2}, which assigns equal weights for all pedestrians and their features, performs worse than social-aware information selection, because the simple implementation does not pay sufficient attention to important neighbors and important trajectories extracted by the features. 
\end{itemize}


\section{Experiments}\label{sec:experiment}

\subsection{Datasets and Metrics}

We evaluate our proposed model on two public pedestrian walking datasets, ETH\cite{pellegrini2009you} and UCY\cite{lerner2007crowds}, which contain rich social interactions. These two datasets contain 5 crowd sets, including ETH-univ, ETH-hotel, UCY-zara01, UCY-zara02 and UCY-univ. There are 1536 pedestrians in total with thousands of non-linear trajectories. We evaluate our model on these 5 datasets. We follow the leave-one-out evaluation
methodology in \cite{gupta2018social}.

There are two types of metrics for evaluating the performance of trajectory prediction, including the Mean Average Displacement (MAD) error and Final Average Displacement (FAD) error\cite{pellegrini2009you} in meters.
\begin{itemize}
\item MAD: Mean Euclidean distance between ground truth and predict points of all predicted time steps.
\item FAD: Euclidean distance between ground truth and predicted point of the last frame.
\end{itemize}
The interval of trajectory sequences is set to 0.4 seconds. We take 8 ground truth positions as observation, and predict the trajectories of following 12 time steps, which follows the setting of \cite{pellegrini2009you,alahi2016social,gupta2018social}.
\subsection{Implementation Details}

\begin{figure}[ht!]
\begin{center}

\includegraphics[width=1\linewidth]{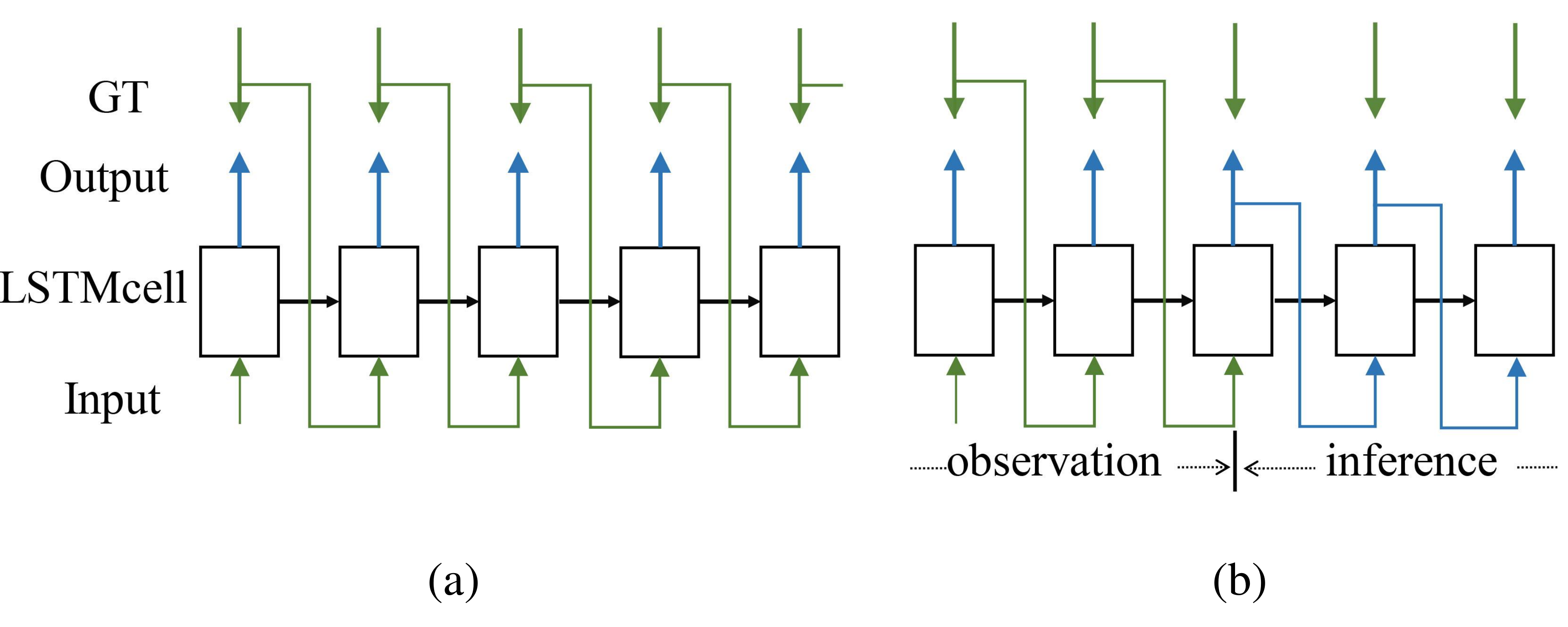}
\end{center}
\caption{Two kinds of teaching mode. (a) Single-step mode. Current ground truth (GT) annotation is given to the next time step as input. (b) Multi-step mode, where the current output is used as the input of next time step at inference stage. }
\vspace{-0.1in}
\label{fig:teach}
\end{figure}

\begin{table*}[t!]
\small
\begin{center}
\begin{tabular}{c|ccccc|c|c|c|c|c|c}
\hline %
\multirow{1}{*}{Variant} &\multicolumn{5}{|c|}{Components} & \multicolumn{6}{|c}{Performance (MAD/FAD)} \\
\cline{2-12}
ID &MG &PA &NS &L &C/P &ETH-univ  &ETH-hotel &UCY-zara01 &UCY-zara02 &UCY-univ &AVG\\
\hline
\hline
1&- &- &2 &1 &C &0.76/1.64 &0.37/0.77 &0.44/0.97 &0.37/0.82 &0.55/1.21 &0.50/1.08\\
2&- &- &10 &1 &C &0.79/1.71 &0.41/0.89 &0.47/1.07 &0.38/0.85 &0.56/1.27 &0.52/1.16\\
3&\checkmark &- &10&1 &C&0.69/1.35 &0.40/0.83 &0.43/0.95 &0.36/0.80 &0.53/1.16 &0.48/1.02\\
4&-&\checkmark &10&1 &C&0.67/1.43 &0.39/0.81 &0.47/1.09 &0.36/0.80 &0.54/1.19 &0.49/1.06\\
5&\checkmark &\checkmark&10 &1 &C&0.64/1.28 &0.39/0.78 &0.42/0.92 &0.34/0.74 &0.52/1.13 &0.46/0.97\\
6&\checkmark &\checkmark&2 &1 &C&0.71/1.45 &0.37/0.75 &0.43/0.93 &0.40/0.97 &0.54/1.21 &0.49/1.06\\

7&\checkmark &\checkmark&10 &2 &C&\textbf{0.63}/\textbf{1.25} &\textbf{0.37/0.74} &\textbf{0.41/0.90} &\textbf{0.32/0.70} &\textbf{0.51}/1.10 &\textbf{0.45}/\textbf{0.94}\\

8&\checkmark &\checkmark&10 &3 &C &0.64/1.27 &0.38/0.75 &0.42/0.91 &0.32/0.71 &0.51/\textbf{1.10} &0.45/0.95\\

9&\checkmark &\checkmark&10 &1 &P &0.71/1.42 &0.39/0.87 &0.47/1.05 &0.35/0.78 &0.53/1.16 &0.49/1.06\\

\hline
\end{tabular}
\end{center}
\caption{Ablation Study on SR-LSTM. \textbf{MG} denotes introducing the motion gate, \textbf{PA} denotes the pedestrian-wise attention layer. \textbf{NS} denotes the neighborhood size in meters, the value of 10 and 2 respectively give a neighborhood region of $20 \times 20$ and $ 4\times 4$. \textbf{L} is the refinement iterations. \textbf{C/\/P} denotes that we use current or previous hidden states to perform the refinement. Variant 1,2 perform the simple states refinement without any feature selection (Eq.\ref{eq:Mp}).}
\label{table:ablation}
\end{table*}
We use single layer MLP to embed the input vectors to 32 dimension, and set the dimension of LSTM hidden state as 64.
A sliding time window with a length of 20 and a stride size of 1 is used to get the training samples.
All trajectory segments in the same time window are regarded as a mini-batch, as they are processed in parallel. We set the size of mini-batch to 8 during the training stage. We use the single-step mode for training (Fig. \ref{fig:teach} (a)), and multi-step mode for validating and testing (Fig. \ref{fig:teach} (b)). Adam optimizer is adopted to train models in 300 epochs, with an initial learning rate of 0.001. For training the model with multiple states refinement layers, we fixed all basic parameters and only learn the parameters of the additional refinement layer.

\subsection{Ablation Study}
\subsubsection{Data pre-processing}

We detail our pretreatment as follows:
\begin{itemize}
\item Relative position or normalized absolute position (Rela/Nabs): Two alternative ways of pre-processing, differentiate the trajectory as relative location offset or shift the origin to the latest observed time step.
\item ETH-univ frame rate issue (EUf): For ETH-Univ scenario, the original video from \cite{ucydatasest} is an accelerated version. We treat every 6 frames as 0.4s, rather than 10 frames in \cite{gupta2018social}.
\item Random Rotation (RR):
For one mini-batch, random rotation is employed for data augmentation.

\end{itemize}
Table \ref{table:preprocess} shows the results of different data pre-processings on V-LSTM, which shows that:
1) Normalized absolute location is superior to relative position in our trials.
2) Correction of the ETH-Univ frame rate significantly promotes for about 12.7/9(\%). 
3) Random rotation is also helpful for reducing overfitting. We adopt data pre-processing configuration of ID4 and use the result of which as baseline.
\subsubsection{Component analysis}
\par
We analyze the components of the proposed model, including the Motion Gate (MG) (Eq. \ref{eq:gm}), the Pedestrian-wise Attention layer (PA) (Eq. \ref{eq:attention}), and the number of refinement layers (L). As we consider the finite neighborhood region, we also take the region size as a variable denoted as Neighborhood Size (NS) in meters. To testify the efficiency of the utilize of current neighboring feature, we also consider using Current or Pervious(C/\/P) hidden states in Eq. \ref{eq:M}. For all variants without PA, we divide the number of neighbors on message passing term for normalization. The quantitative results of different model variants are reported in Table \ref{table:ablation}.  

\textbf{Simple states refinement.}
Performing the simple states refinement (Eq.\ref{eq:Mp2}) without any feature selection and considering the neighborhood size of 2 meters (Variant 1) outperforms V-LSTM by 6.4/6.8(\%), as the human interaction is involved through the states refinement module. But the model with neighborhood size of 10 meters results in slight changes (1.4/-0.2(\%)). The effect of neighborhood size is summarized in following paragraph.

\textbf{Neighborhood size.}
We test two value of neighborhood size, 2 and 10, the effect of which are summarized: 1) Simple states refinement model with the equal treatment of all pedestrians within 10 meters (Variant 2), where useless features from far neighbors are still considered for message passing, causes performance deterioration of 5.6/7.5(\%) relative to the same model with the neighborhood size of 2 meters. 
2) With the proposed information selection mechanism, considering larger neighborhood size is generally better (Variant 5 vs 6). Therefore, our SR-LSTM could take advantage of useful information from farther neighbors. 

\textbf{Information selection.}
With neighborhood size fixed as 10 meter, only introducing the motion gate (Variant 3) or pedestrian-wise attention (Variant 4) is resultful, which respectively improves the performance by 7.8/12.2(\%) and 6.7/8.3(\%). Utilization of both these two components (Variant 5) achieves the improvement of 11.8/16.4/(\%), which demonstrates the effectiveness of our information selection mechanism.
When neighborhood size is set to 2 meters, adding motion gate and pedestrian-wise attention (Variant 6) still outperforms the simple refinement model (Variant 1) on average.\par

\textbf{States refinement from current states.}
Utilization of the current states (Variant 5) outperforms the one using the previous states (Variant 9) by 6/8.3(\%), which demonstrates the importance of latest features of neighbors.

\textbf{Refinement iterations.}
Employing the second states refinement layer (Variant 7) performs consistently better than only refine the states once (Variant 5) by 2.8/3(\%). While the third layer introduced could not bring further promotion. It may suggest that the choice of two refinement iterations is the appropriate for this task.

\subsection{Comparison with Existing Works}
\begin{table*}[t!]
\small
\begin{center}
\begin{tabular}{c|cccc|c|c|c|c|c|c}
\hline %
\multirow{2}{*}{Method} &\multicolumn{4}{|c|}{\multirow{2}{*}{Notes}} & \multicolumn{6}{|c}{Performance (MAD/FAD)} \\
\cline{6-11}
& & & & &ETH-univ  &ETH-hotel &UCY-zara01 &UCY-zara02 &UCY-univ &AVG\\
\hline
\hline
V-LSTM* &\multicolumn{4}{c|}{-} &1.09/2.41 &0.86/1.91 &0.41/0.88 &0.52/1.11 &0.61/1.31 &0.7/1.52\\
SGAN* &\multicolumn{4}{c|}{20 samples} &0.81/1.52 &0.72/1.61 &0.34/0.69 &0.42/0.84 &0.60/1.26 &0.58/1.18\\
Sophie* &\multicolumn{4}{c|}{20 samples+scene} &0.70/1.43 &0.76/1.67 &\textbf{0.30/0.63} &0.38/0.78 &0.54/1.24 &0.54/1.15\\
S-LSTM\_1 &\multicolumn{4}{c|}{NS=2, grid: 4$\times$4} &0.70/1.40 &\textbf{0.37/0.73} &0.49/1.15 &0.39/0.89 &0.60/1.32 &0.51/1.10\\
S-LSTM\_2 &\multicolumn{4}{c|}{NS=10 grid: 4$\times$4} &0.77/1.60 &0.38/0.80 &0.51/1.19 &0.39/0.89 &0.58/1.28 &0.53/1.15\\
V-LSTM    &\multicolumn{4}{c|}{-}&0.83/1.77 &0.41/0.80 &0.49/1.15 &0.37/0.85 &0.56/1.22 &0.53/1.16\\
\hline
SR-LSTM\_1&\multicolumn{4}{c|}{ID 6 in Tab.\ref{table:ablation}}&0.64/1.28 &0.39/0.78 &0.42/0.92 &0.34/0.74 &0.52/1.13 &0.46/0.97 \\

SR-LSTM\_2&\multicolumn{4}{c|} {ID 7 in Tab.\ref{table:ablation}}&\textbf{0.63}/\textbf{1.25} &0.37/0.74 &0.41/0.90 &\textbf{0.32/0.70} &\textbf{0.51/1.10} &\textbf{0.45}/\textbf{0.94}\\
\hline
\end{tabular}
\end{center}
\caption{Comparison with several baselines models. \textbf{NS} denotes the neighborhood size in meters. The results of methods marked with * are directly obtained from \cite{gupta2018social,sadeghian2018sophie}. }
\label{table:quatitative}
\end{table*}

We compare our model with several recent existing works: (1) Social-LSTM\cite{alahi2016social}: A cubic tensor is used in this approach to gather the social information. The recommended neighborhood size is 32 pixels in image space, we choose it as 2 and 10 meters respectively referred as S-LSTM\_1 and S-LSTM\_2.
(2) SGAN\cite{gupta2018social}: A multimodal method to retrieve multiple possible future paths.
(3) Sophie\cite{sadeghian2018sophie}: An improved multimodal method which introduces the attention on social relationship and physical acceptability.

The results are shown in Table \ref{table:quatitative}. All of methods are under the same dataset setting and evaluation methodology.
Note that SGAN and Sophie report the results that best match groundtruth in 20 samples, the other methods only produce one prediction; Sophie also requires the scene image.

\textbf{V-LSTM vs V-LSTM*.}
V-LSTM models implemented by ourselves in Table \ref{table:preprocess} could not completely match the result of V-LSTM*, which is reported in \cite{gupta2018social}. This is possibly due to the deviation on hyper-parameters, data organization, or the teaching mode. In addition, we try our best to search for better data reprocessing which results in a considerable promotion.

\textbf{SR-LSTM vs others.}
By the captured multimodality, SGAN and Sophie improve significantly in comparison with V-LSTM*. But SGAN could not outperform V-LSTM* with only a single sample\cite{gupta2018social}. Our best model increases the performance relative to V-LSTM for 15.4/18.8(\%), with only a single prediction.

S-LSTM\_1 outperforms V-LSTM but still has higher prediction error than our approach, because it only takes advantage of previous hidden states of local neighbors. In addition, S-LSTM is not able to take advantage of the far neighbors according to the results of S-LSTM\_2. Our SR-LSTM makes it possible to consider far neighbors and utilize their current states to refine each other.
\subsection{Qualitative Results}

\begin{figure}[h]
\centering
\subfigure[]{
\includegraphics[width=0.77\linewidth]{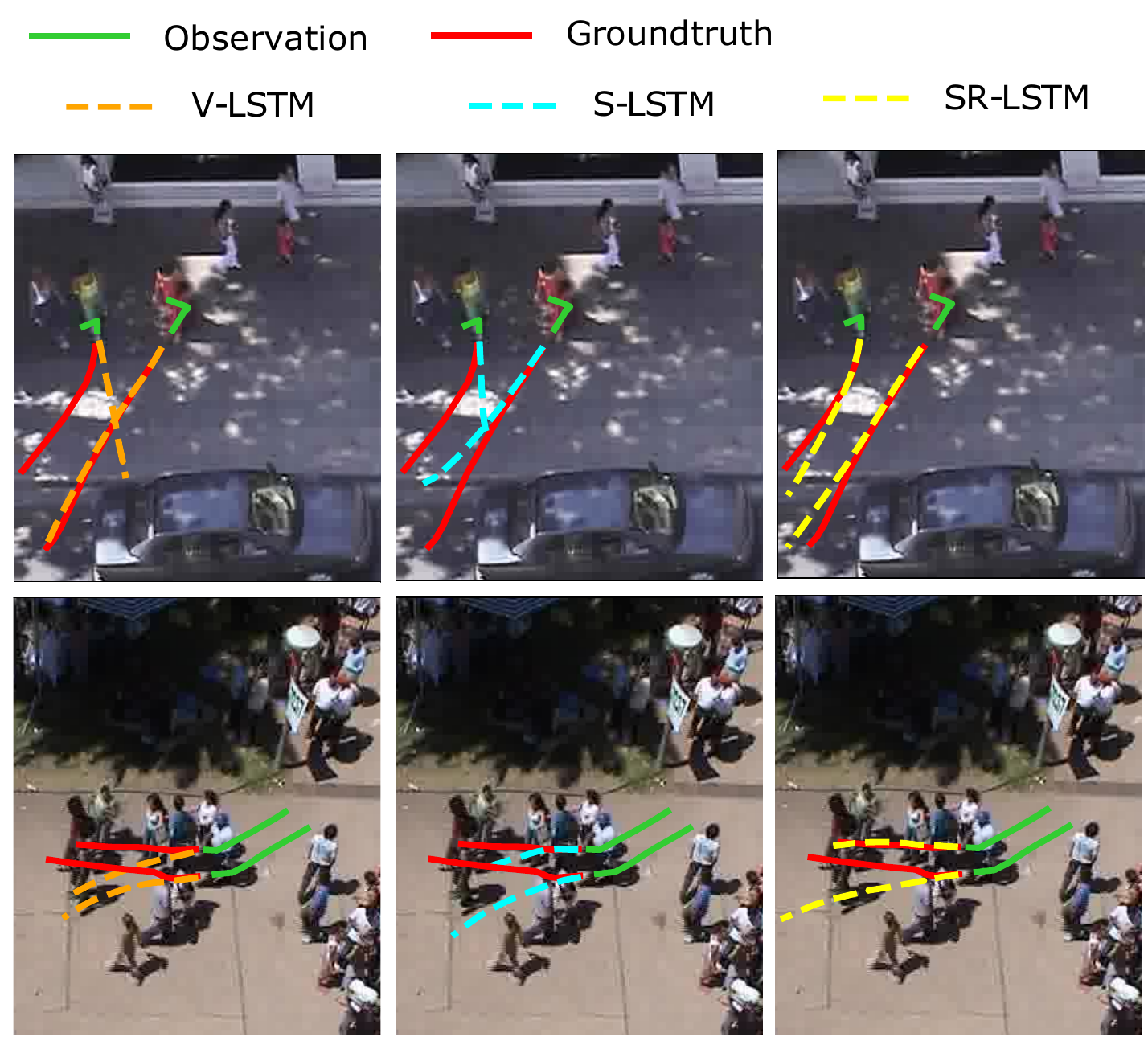}\label{fig:current}}
\subfigure[]{
\includegraphics[width=0.87\linewidth]{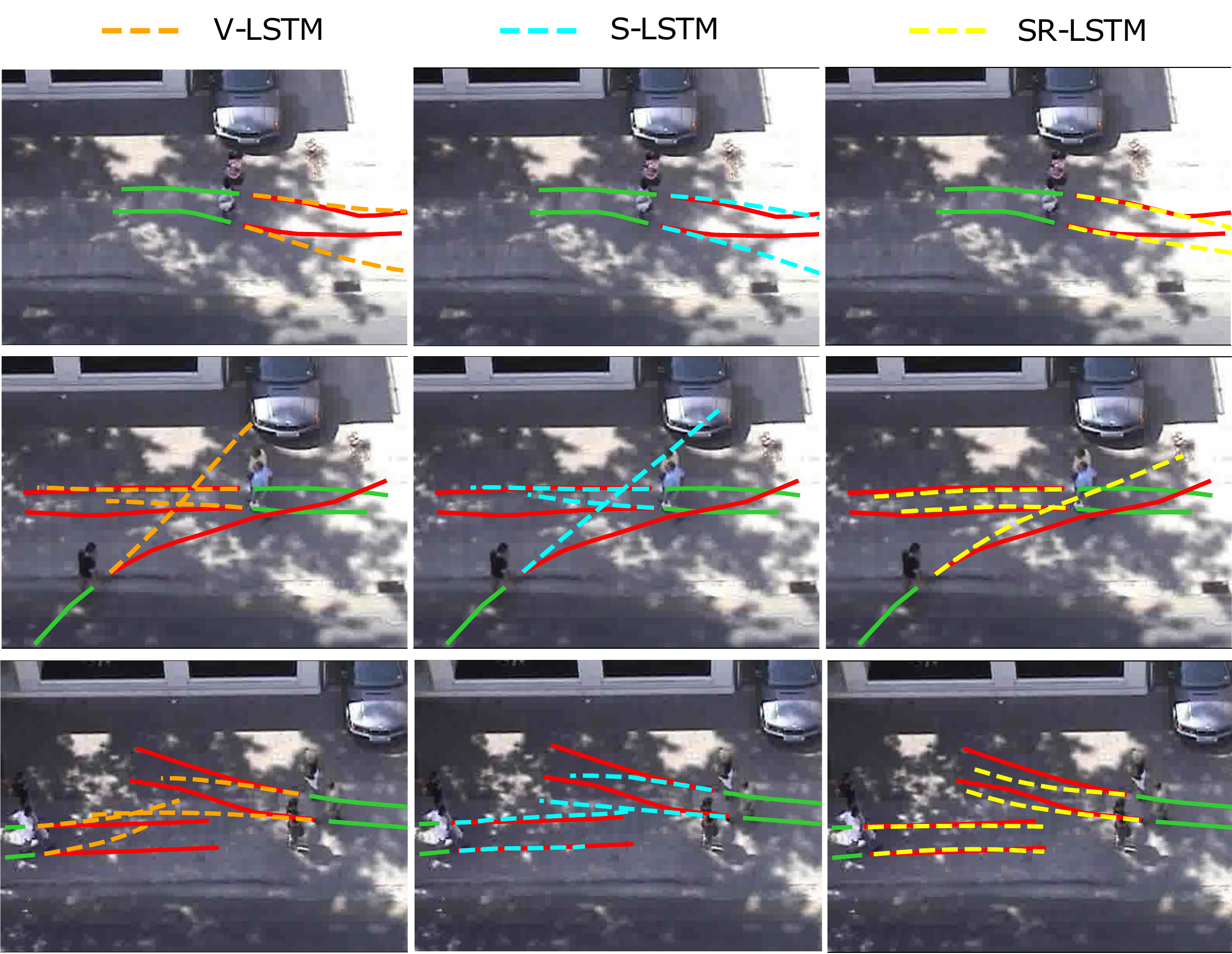}\label{fig:social}}

\caption{Illustration of the prediction trajectories. (a). In SR-LSTM, current states of pedestrians can timely refine each other, particularly in the case where pedestrians change their intentions. (b). SR-LSTM are able to implicitly explain for common social behaviors, which gives moderate future predictions and relatively low errors.}

\end{figure}

\textbf{Feature refinement from current states.} Benefiting from our states refinement module, SR-LSTM is able to take advantage from the current neighboring states. Fig. \ref{fig:current} shows examples in which pedestrians' walking direction have suddenly changed before few time steps. V-LSTM (first column) does not consider the interaction and results in large error. S-LSTM (S-LSTM\_1 in Table \ref{table:quatitative}, second column) utilizes the previous neighboring LSTM states, but is still insensitive to these cases. Our SR-LSTM (SR-LSTM\_2 in Table \ref{table:quatitative}, third column) refines the current LSTM states through message passing, which can timely capture changes of the others' intention and make suitable adjustment.

\textbf{Social behaviors.} SR-LSTM can moderately explain implicit social behaviors. In Fig.\ref{fig:social}, we illustrate three cases, consistent group walking, collision avoidance and group avoidance. In V-LSTM, pedestrians are walking in their own. S-LSTM performs weaker to model pedestrian interactions and ignores the potential effect from far neighbors. Our SR-LSTM shows pretty ability to make appropriate prediction towards social interaction.  \par

\subsection{Social-aware Information Selection}

\begin{figure}[ht!]
\centering
\includegraphics[width=1\linewidth]{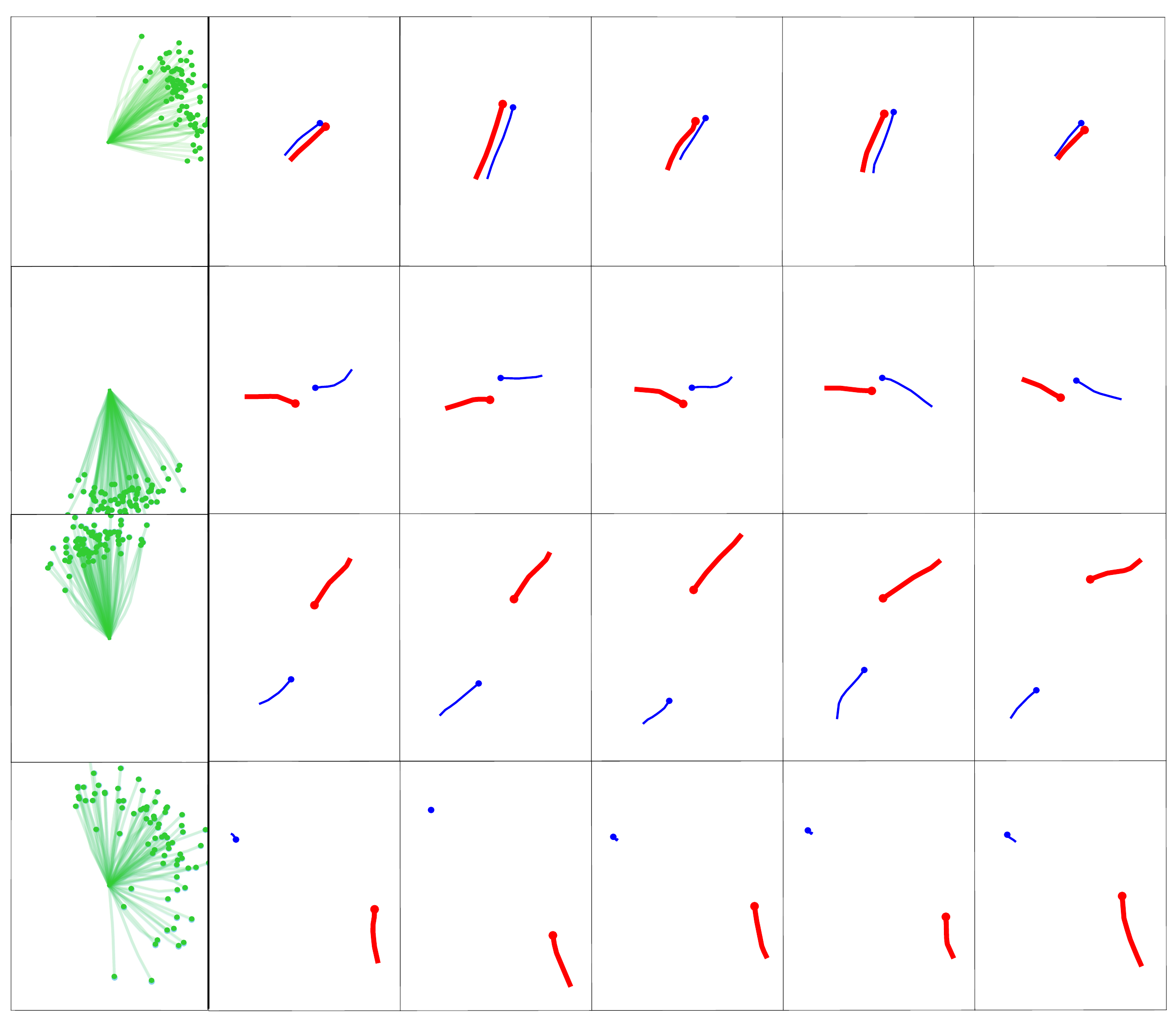}
\caption {Selected feature patterns by motion gate. Each row is related to a hidden neuron (feature) of LSTM. Column 1: Activation trajectory pattern of the hidden feature. Column 2-6: Pairwise trajectory examples (end with solid dots) having high activation to the motion gate. Prediction for the pedestrian in red is mostly sensitive to the other's potential trajectories showed in first column, which are selected by our motion gate.}\label{fig:gatepattern}
\end{figure}

\textbf{Motion gate.} When predicting the position of pedestrian $i$, motion gate acted on the hidden features of his/her neighbor $j$ is calculated based on the pairwise features between pedestrian $i$ and $j$ (Eq.\ref{eq:gm}). Fig.\ref{fig:gatepattern} shows how motion gate selects the features, where each row is related to a certain dimension of hidden feature.

In Fig.\ref{fig:gatepattern}, the first column shows the trajectory patterns captured by hidden features started from origin and ended at the dots, which are extracted in similar way as Fig.\ref{fig:motivation2}(a).
The motion gate for a feature considers pairwise input trajectories with similar configurations. Some examples for high response of the gate are shown in the other columns of Fig.\ref{fig:gatepattern}.
In these pairwise trajectory samples, the red and blue ones are respectively the trajectories of pedestrian $i$ and $j$, and the time step we calculate the motion gate are shown with dots (where the trajectory ends). These pairwise samples are extracted by searching from database with highest  activation for the motion gate neuron. High response of gate means that the corresponding feature is selected.

As shown in Fig.\ref{fig:gatepattern},
a gate for the same feature is responsible for roughly similar interaction conditions. When predicting the trajectory of pedestrian $i$ (red), our motion gate attentively select features of pedestrian $j$ (blue). These selected features shown in first column represent the potential trajectories that the pedestrian $j$ might cause future interaction with the pedestrian $i$.

We explain effects of four gate elements in each row of Fig.\ref{fig:gatepattern}:
1) Row 1: The trajectory pairs are very close and are walking together. The selected hidden feature follows the walking direction. 
2) Row 2: The trajectories are somewhat close but walking in opposite direction. The pedestrian $i$ in red cares about whether the other will walk towards him/her.
3) Row 3: This case is similar to row 2. This gate element considers more distant neighbor walking in opposite direction.
4) Row 4: The neighbor in blue is static, the selected hidden feature shows that pedestrian $i$ in red potentially pay attention on this stationary neighbor in case he is about to walk towards him/her.

\begin{figure}[ht!]
\centering
\includegraphics[width=1\linewidth]{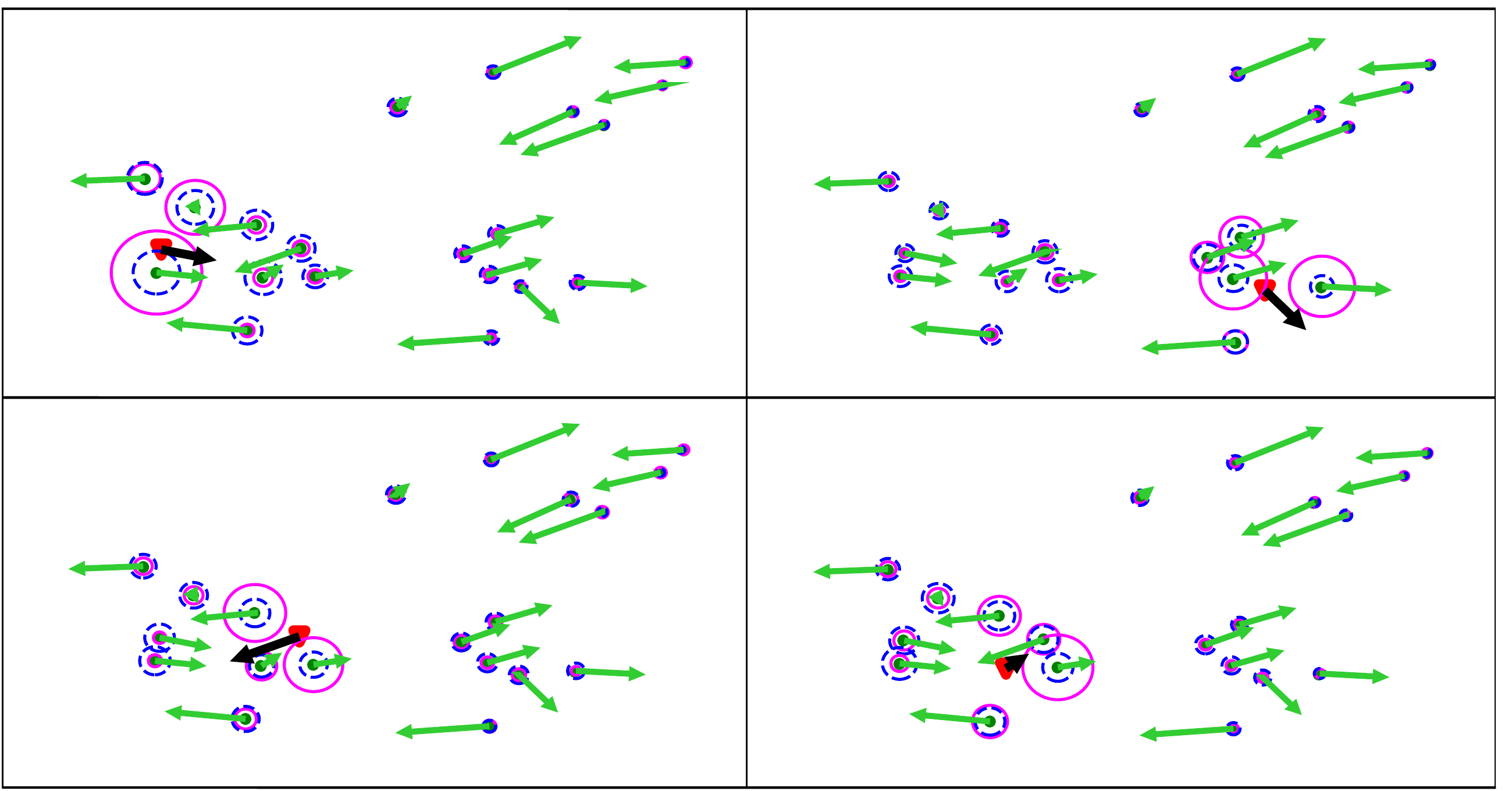}
\caption{Illustration of the pedestrian-wise attention. Circle in magenta represents the attention in first round states refinement, the dashed circle represents for the attention in the second refinement. Larger circle corresponds to higher attention. Red triangle represents the target pedestrian for trajectory prediction, and green ones are his/her neighbors, the arrows on each of them represent their walking directions.} \label{fig:attnpattern}
\end{figure}

\textbf{Pedestrian-wise attention.}
We illustrate some examples of the pedestrian-wise attention expected by our SR-LSTM in Fig.\ref{fig:attnpattern}. It shows that 1) dominant attention is paid to the close neighbors, while the others also take slight attention, 2) the attention given by the first refinement layer often largely focuses on the close neighbors, and the second refinement tends to strengthen the effect of farther neighbors with group behavior or may influence the pedestrian in longer time range.

\section{Conclusion}
In this paper, we propose a states refinement module for LSTM network to address the the problem of joint trajectory prediction for pedestrians in the crowd. Our states refinement module treats LSTM as feature extractor, which adaptively refines current features of all pedestrians based on a message passing mechanism. In addition, we introduce a social-aware information selection mechanism consisting of an element-wise motion gate and a pedestrian-wise attention, to select useful features of each neighbor. The states refinement module with information selection outperforms the state-of-the-art approaches.
{\small
\bibliographystyle{ieee}
\bibliography{egbib}
}

\end{document}